\documentclass[runningheads]{llncs}
\usepackage{graphicx}
\usepackage{amsmath,amssymb}
\usepackage{color}
\usepackage{orcidlink}
\usepackage{cite}
\usepackage{booktabs}
\usepackage{multirow}
\usepackage{longtable}

\begin{document}

\title{SIM2E: Benchmarking the Group\\ Equivariant Capability of Correspondence\\ Matching Algorithms}

\titlerunning{SIM2E Benchmark}

\author{Shuai Su\orcidlink{0000-0001-6144-8923} \and
Zhongkai Zhao\orcidlink{0000-0003-2365-9898} \and
Yixin Fei\orcidlink{0000-0001-9852-0415} \and
Shuda Li\orcidlink{0000-0003-4216-8074} \and
\\
Qijun Chen\orcidlink{0000-0001-5644-1188} \and
Rui Fan\orcidlink{0000-0003-2593-6596}}
\authorrunning{S. Su \textit{et al}.}
\institute{Tongji University, Shanghai 201804, China.\\
\email{\{sushuai, kanez, amyfei, qjchen\}@tongji.edu.cn, shuda.dexter.li@gmail.com, rui.fan@ieee.org}
}

\maketitle

\begin{abstract}
Correspondence matching is a fundamental problem in computer vision and robotics applications. Solving correspondence matching problems using neural networks has been on the rise recently. Rotation-equivariance and scale-equivariance are both critical in correspondence matching applications. Classical correspondence matching approaches are designed to withstand scaling and rotation transformations. However, the features extracted using convolutional neural networks (CNNs) are only translation-equivariant to a certain extent. Recently, researchers have strived to improve the rotation-equivariance of CNNs based on group theories. Sim(2) is the group of similarity transformations in the 2D plane. This paper presents a specialized dataset dedicated to evaluating sim(2)-equivariant correspondence matching algorithms. We compare the performance of 16 state-of-the-art (SoTA) correspondence matching approaches. The experimental results demonstrate the importance of group equivariant algorithms for correspondence matching on various sim(2) transformation conditions. Since the subpixel accuracy achieved by CNN-based correspondence matching approaches is unsatisfactory, this specific area requires more attention in future works. Our dataset is publicly available at: \href{https://mias.group/SIM2E/}{mias.group/SIM2E}. 

\end{abstract}

\section{Introduction}

\begin{figure}[t!]
\centering
\includegraphics[width=0.99\textwidth]{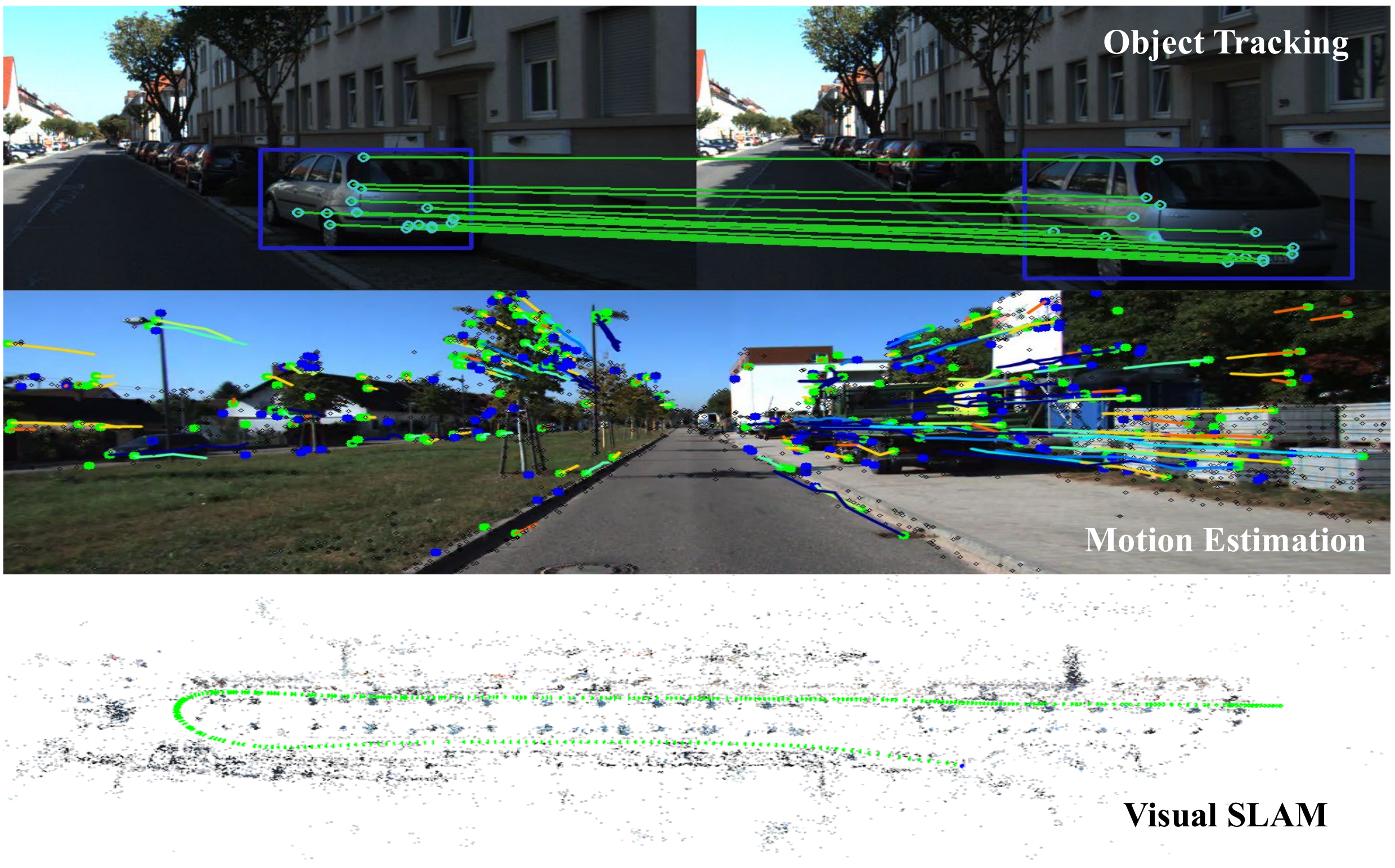}
\caption{Autonomous driving perception tasks involving correspondence matching.}
\label{fig:applications}
\end{figure}

Correspondence matching is a key component in autonomous driving perception tasks, such as object tracking \cite{zhou2009object}, simultaneous localization and mapping (SLAM) \cite{yu2022accurate}, multi-camera online calibration \cite{ling2016high}, 3D geometry reconstruction \cite{fan2018road}, panorama stitching \cite{brown2007automatic}, and camera pose estimation \cite{fan2019road}, as shown in Fig. \ref{fig:applications}. Sim(2) transformation consists of rotation, scaling, and translation. Sim(2)-equivariant correspondence matching is significantly important for autonomous driving, as vehicles often veer abruptly.
Classical algorithms leverage a detector, a descriptor, and a matcher to determine correspondences. The detector and descriptor provide the locations and the descriptions of interest points (point-like features in an image), and the matcher produces the final correspondences. Moravec \textit{et al}. \cite{moravec1977techniques} presented the concept of interest points. Harris \cite{harris1988combined} judges whether the pixel is a corner based on the local image gradient changes. The scale-invariant feature transform (SIFT) \cite{lowe1999object} is a rotation-invariant and scale-invariant algorithm that consists of a detector and a descriptor. The distance among descriptors is computed using cosine distance. The correspondences of a given image pair are determined using the nearest neighbor matching algorithms. As a hand-crafted algorithm, SIFT \cite{lowe1999object} achieves rotation-invariance by computing the main directions of local features (in an image patch of 16x16 pixels.). ASIFT \cite{morel2009asift} aimed to improve the performance of SIFT \cite{lowe1999object} on affine transformation. Oriented FAST \cite{rosten2006machine} and Rotated Binary Robust Independent Elementary Features (BRIEF) \cite{calonder2010brief} (ORB) \cite{rublee2011orb} greatly minimize the trade-off between accuracy and speed and have been widely used in visual SLAM \cite{mur2015orb, mur2017orb, rublee2011orb}.

Deep learning has been applied successfully in numerous computer vision tasks in recent years. LIFT \cite{yi2016lift} is an architecture of learning-based rotation-invariant feature detection and description approach. It consists of three modules: detector, orientation estimator, and descriptor. Similar to SIFT \cite{lowe1999object}, a scale-space pyramid is used to obtain multi-scale correspondence detection results. SuperPoint \cite{detone2018superpoint} is a self-supervised framework for correspondence detection and description. It is a two-stage method: (1) in the first stage, a feature extractor is trained on a synthetic dataset generated by rendering patterns of corners; (2) in the second stage, the descriptor is trained on images from the COCO-dataset \cite{lin2014microsoft} that are augmented by random homography matrices including transformations such as rotation, scaling, and translation. Unlike SuperPoint, D2Net \cite{dusmanu2019d2} is a one-stage approach that jointly detects and describes correspondences. D2Net is trained using the correspondences obtained from large-scale structure from motion (SfM) reconstructions.
R2D2 \cite{revaud2019r2d2} proposes a framework to find more repeatable correspondences, and it can simultaneously estimate the reliability and repeatability of correspondences. DISK \cite{tyszkiewicz2020disk} uses reinforcement learning to realize end-to-end correspondence matching. It performs better at small angle changes but worse than SuperPoint \cite{detone2018superpoint} at large angle changes.

SuperGlue \cite{sarlin2020superglue} achieves superior performance by modeling correspondence matching as a graph matching problem. The inputs of the graph neural network in SuperGlue \cite{sarlin2020superglue} include descriptors, positions, and scores of keypoints. The Sinkhorn algorithm \cite{cuturi2013sinkhorn} is utilized to solve the optimal-transport problem. However, the graph edges in SuperGlue exponentially grow as the number of correspondences increases. SGMNet \cite{chen2021learning} uses a seeded graph to reduce computation and memory costs significantly. LoFTR \cite{sun2021loftr} is a detector-free and end-to-end architecture. It uses convolutional neural networks (CNN) as feature extractors and a coarse-to-fine strategy to obtain more accurate pixel-level results. Similar to SuperGlue, it also fuses descriptors with the position information. Unlike LoFTR, another end-to-end correspondence matching network, referred to as MatchFormer \cite{wang2022matchformer}, uses a hierarchical extract-and-match transformer. It is demonstrated that the correspondence matching operation can also be conducted in the encoder. RoRD \cite{parihar2021rord} uses orthographic view generation to improve correspondence matching by increasing the visual overlap using orthographic projection. It also shows that rotation invariance can be improved by augmenting the training dataset with random rotation, scaling, and perspective transformations.

Group-equivariant convolutional neural networks (G-CNN) are equivariant under a specific transformation (\textit{e.g.}, rotation, translation, \textit{etc}.) which can also be represented by a special group. Researchers have designed G-CNNs using different mathematical approximations. Cohen \textit{et al}. \cite{cohen2016group} proposed the first G-CNN. Li \textit{et al}. \cite{li2018deep} use the cyclic replacement to achieve P4-group equivariance. Cohen \textit{et al}. \cite{cohen2018spherical} use the Fast Fourier Transform (FFT) to approximate the integral of a group. E2-CNN \cite{weiler2019general} is a general G-CNN framework that analyzes and models the orientation and symmetry of images. 
GIFT \cite{liu2019gift} is a rotation-equivariant and scaling-equivariant descriptor based on G-CNN. It uses E2-CNN \cite{weiler2019general} rather than conventional CNNs to describe local visual features. On the other hand, SEKD \cite{lee2022self} is a group-equivariant correspondence detector based on G-CNN, which greatly improves the performance of rotation-equivariant correspondence matching. ReF \cite{peri2022ref} is a rotation-equivariant correspondence detection and description framework. It uses a G-CNN to extract group-equivariant feature maps and a group-pooling operation to get rotation-invariant descriptors. SE2-LoFTR \cite{bokman2022case} replaces the feature extractor of LoFTR with E2-CNN, achieving significantly better results on the rotated-HPatches dataset \cite{parihar2021rord}. 
Furthermore, it also mentioned in \cite{bokman2022case} that the the position information is not rotation-equivariant while the descriptor is rotation-equivariant. The methods mentioned above only consider the equivariance of local features. Unfortunately, the equivariance of position information is rarely discussed. Cieslewski \textit{et al}. \cite{cieslewski2018matching} presented an algorithm to match correspondences without descriptors, namely, only position information is used. This algorithm is evaluated on the KITTI \cite{geiger2012we} dataset (containing relatively ideal scenarios), demonstrating robust performance even without descriptors. 
Similar to \cite{cieslewski2018matching}, ZZ-Net \cite{bokman2021zz} is an algorithm for matching two 2D point clouds. It demonstrates that correspondence matching without descriptors can work in rotation-only conditions. Therefore, the current research on the equivariance of position information needs to be further expanded.

\section{SIM2E Dataset}
\subsection{Data Collection and Augmentation}
To ensure the pixel-level accuracy of correspondence matching ground truth, we scrape frames from online time-lapse videos. The cameras used to capture such time-lapse videos are fixed. Our SIM2E dataset contains many challenging scenarios, such as moving clouds in the sky and changing illumination conditions. We choose the first frame of each video as the reference image and use the rest of the frames as target (query) images. We also publish our data augmentation code so that interested readers can conduct sim(2) transformations on our dataset according to their own needs.

\subsection{SIM2E-SO2S, SIM2E-Sim2S, and SIM2E-PersS}

The rotation and scaling operations produce many black backgrounds. To increase the difficulty of correspondence matching, we generate synthetic backgrounds to fill these black areas. Our dataset is split into three subsets: SIM2E-SO2S, SIM2E-Sim2S, and SIM2E-PersS.
\begin{itemize}
\item \textbf{SIM2E-SO2S (Rotation and Synthetic Background)}: The target images are rotated by random angles between 0$^\circ$ and 360$^\circ$. Scaling is not applied. 
\item \textbf{SIM2E-Sim2S (Rotation, Scaling, Translation and Synthetic Background)}: The target images are rotated by random angles between 0$^\circ$ and 360$^\circ$. Random scaling ranging between 0.4 and 1, and random translation ranging between 0 and 0.2 are also applied.
\item \textbf{SIM2E-PersS: (Perspective Transformation and Synthetic Background)}: Random perspective transformations are applied to the target images, where the perspective parameters (the two elements on the 3rd row, the 1st and 2nd columns of the homography matrix, respectively) are random values between -0.0008 and 0.0008. The shear angle is randomly set to $[-10^\circ, 10^\circ]$. The target images are rotated by random angles between 0$^\circ$ and 360$^\circ$. Random scaling ranging between 0.4 and 1 is applied. Random translation ranging between 0 and 0.2 is applied.
\end{itemize}

\begin{figure}[t!]
\centering
\includegraphics[width=0.99\textwidth]{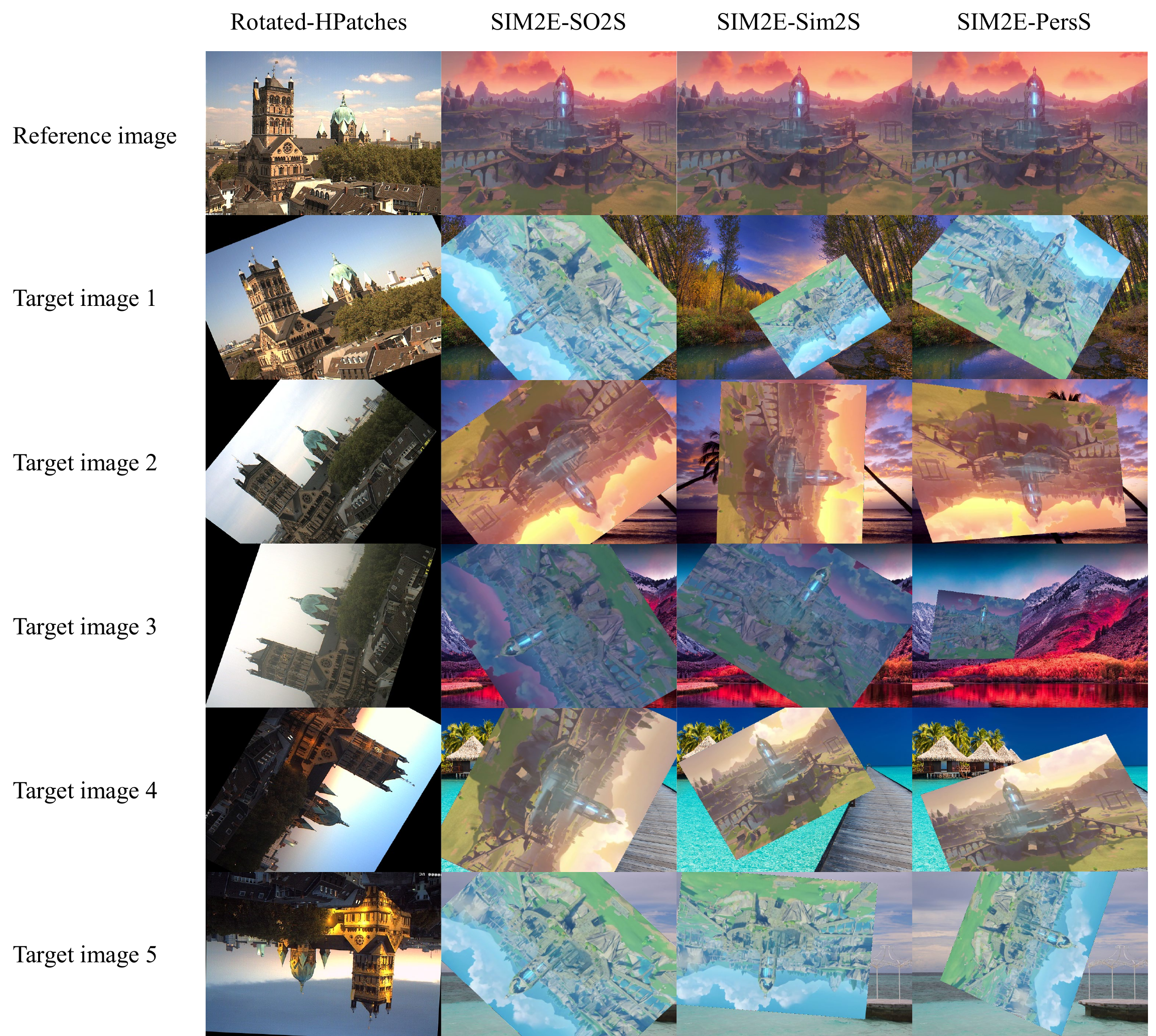}
\caption{Rotated-HPatches dataset and the three sub-sets of our created SIM2E dataset.}
\label{fig:dataset}
\end{figure}

\subsection{Comparison with Other Public Correspondence Matching Datasets}
As shown in Table. \ref{tab:dataset}, the existing datasets for correspondence matching algorithm evaluation can be grouped into two types: 3D scenes \cite{sattler2012image, sattler2018benchmarking, dai2017ScanNet, li2018MegaDepth, taira2018inloc, wang2020tartanair} and planar scenes \cite{balntas2017hpatches, parihar2021rord}. 

\begin{table}[t!]
    \centering
        \caption{Comparison between our SIM2E dataset and other public datasets.}
    \label{tab:dataset}
    \begin{tabular}
        {l|c|c|c|c|c}
        \toprule
        Dataset          &  Type & Illumination Change & Rotation & Scaling  &  Dataset Size    \\
        \hline
        AachenDayNight \cite{sattler2012image}   &  3D   &     significant    &  small  &  medium &  large        \\
        ScanNet \cite{dai2017ScanNet}            &  3D   &     slight    &  small  &  small  &  large        \\
        MegaDepth \cite{li2018MegaDepth}         &  3D   &     slight    &  small  &  large  &  large        \\
        Inloc \cite{taira2018inloc}              &  3D   &    medium    &  small  &  medium &  large        \\
        TartanAir \cite{wang2020tartanair}       &  3D   &  very significant  &  small  &  medium &  large        \\
        \hline
        Hpatches \cite{balntas2017hpatches}      & plane &     significant    &  small  &  small  &  small        \\
        Rotated-HPatches \cite{parihar2021rord}  & plane &     significant    &  large  &  medium &  small        \\
        \hline
        \textbf{SIM2E (ours)}                              & plane &  very significant  &  large  &  large  &  small        \\
        \bottomrule
    \end{tabular}
\end{table}

Aachen Day-Night \cite{sattler2012image} is a public dataset designed to evaluate the performance of outdoor visual localization algorithms in changing illumination conditions (day-time and night-time). The dataset contains a scenario where images were taken with a hand-held camera at different times of the day. It is widely used to evaluate the performance of correspondence matching algorithms, especially when the illumination change is significant. 

ScanNet \cite{dai2017ScanNet} is a large-scale real-world dataset containing 2.5M RGB-D images (1513 scans acquired in 707 different places, such as offices, apartments, and bathrooms). All the scans are annotated with estimated calibration parameters, camera poses, reconstructed 3D surfaces, textured meshes, dense object-level semantic segmentations, and aligned computer-aided design (CAD) models.

MegaDepth \cite{li2018MegaDepth} is a large-scale dataset for the evaluation of depth estimation and/or correspondence matching algorithms. It uses SfM and multi-view stereo (MVS) techniques to acquire 3D point clouds, which can then be used to train and evaluate single-view depth estimation and/or correspondence matching networks. However, the 3D point clouds generated using SfM/MVS in the MegaDepth dataset are not sufficiently accurate and dense.

Compared to the Aachen Day-Night \cite{sattler2012image} and MegaDepth \cite{li2018MegaDepth} datasets which were created in outdoor scenarios, the Inloc \cite{taira2018inloc} dataset focuses on indoor localization problems. The Inloc dataset consists of a database of RGB-D images, geometrically registered to the floor maps and augmented with a separate set of RGB target images (annotated with manually verified ground-truth 6DoF camera poses in the global coordinate system of the 3D map). 

Unlike the aforementioned datasets that are relatively ideal in terms of either motion or illumination conditions, TartanAir \cite{wang2020tartanair}, a synthetic dataset used to evaluate visual SLAM algorithms, is collected using a photo-realistic simulator (with the presence of moving objects, changing illumination and weather conditions). Such a more challenging dataset fills the gap between synthetic and real-world datasets. 

Hpatches \cite{balntas2017hpatches} are created using other public datasets. It can be split into two subsets: illumination and viewpoint, which are two crucial aspects of correspondence matching. It can also be split into three subsets: EASY, HARD, and TOUGH, according to the sizes of the overlapping areas between reference and target images. Randomly rotating the target images in the Hpatches \cite{balntas2017hpatches} dataset produces a new dataset, referred to as Rotated-HPatches \cite{parihar2021rord}. 

Similar to the Hpatches and Rotated-HPatches datasets, our SIM2E dataset provides accurate subpixel correspondence matching ground truth. On the other hand, the illumination, rotation, and scaling changes are significant in our dataset. Therefore, compared to other existing public datasets, our SIM2E dataset can be used to evaluate the sim(2)-equivariant capability of correspondence matching algorithms more comprehensively. However, the size of the current version of our SIM2E dataset is small. We will therefore increase its size in our future work. 

\begin{figure}[t!]
	\centering
	\includegraphics[width=0.99\textwidth]{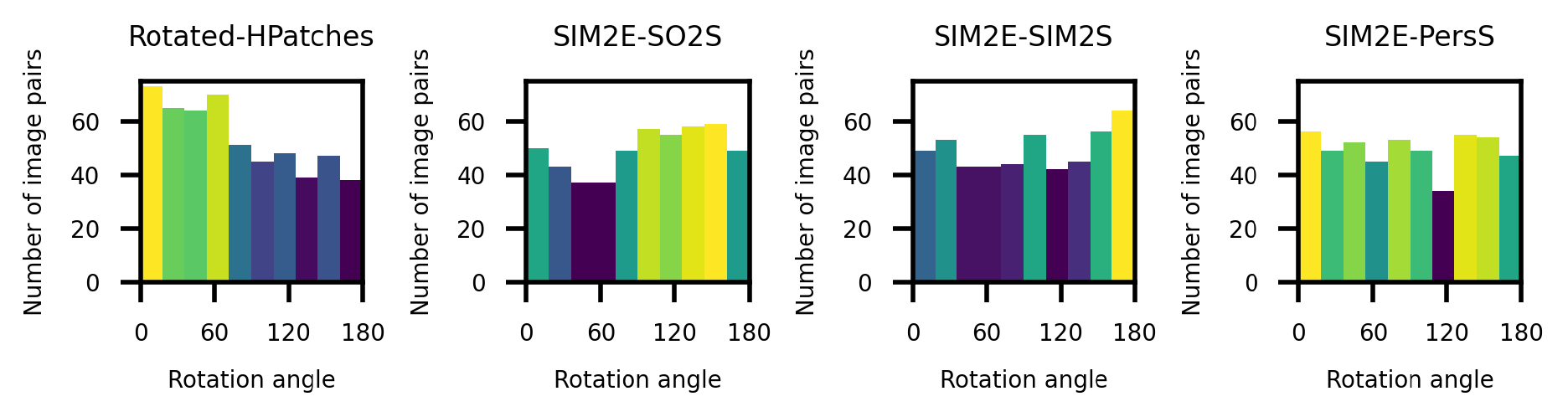}
	\caption{Rotation distributions of the Rotated-HPatches and our SIM2E datasets.}
	\label{fig:distribution}
\end{figure}

The rotation distributions of Rotated-HPatches and our SIM2E subsets are shown in Fig. \ref{fig:distribution}. It can be observed that in the Rotated-HPatches dataset, slight rotations ($\leq60^\circ$) account for a large proportion. In contrast, the three subsets of our SIM2E dataset are uniformly distributed. Most existing learning-based correspondence matching approaches have poor rotation-equivariant capabilities, and their performances are satisfactory only when there are slight rotations. Therefore, our SIM2E dataset can provide more acceptable results when evaluating the rotation-equivariant capability of a given correspondence matching algorithm.
\section{Experiments}
\subsection{Experimental Setup}

The group-equivariant capabilities of six classical and ten learning-based correspondence matching approaches are evaluated on our SIM2E dataset. 

For classical correspondence matching approaches, we use the OpenCV \cite{bradski2000opencv} implementations of AKAZE \cite{alcantarilla2011fast}, BRISK \cite{leutenegger2011brisk}, KAZE \cite{alcantarilla2012kaze}, ORB \cite{rublee2011orb}, FREAK \cite{alahi2012freak}, and SIFT \cite{lowe1999object} in our experiments. All these classical approaches use the nearest neighbor matching algorithm for correspondence matching. The ratio test technique (threshold is set to 0.7) is also used to improve the overall performance.

For learning-based correspondence matching approaches, we use the official weights of each model. These models were trained on different datasets, as detailed below:

\begin{itemize}
\item \textbf{SuperPoint} \cite{detone2018superpoint} is trained on the MS-COCO \cite{lin2014microsoft} dataset, a large-scale dataset for object detection and segmentation.

\item \textbf{R2D2} \cite{revaud2019r2d2} is trained on the Aachen Day-Night \cite{sattler2012image} dataset and a retrieval dataset \cite{radenovic2018revisiting}. 

\item \textbf{ALIKE} \cite{zhao2022alike} is trained on the MegaDepth \cite{li2018MegaDepth} dataset.

\item \textbf{GIFT} \cite{liu2019gift} is trained on the MS-COCO \cite{lin2014microsoft} dataset and finetuned on the GL3D \cite{shen2018matchable} dataset (consisting of indoor and outdoor scenes).

\item \textbf{RoRD} \cite{parihar2021rord} is trained on the PhotoTourism \cite{snavely2006photo} dataset, where the 3D structures of scenes are obtained using SfM.

\item \textbf{SuperGlue} \cite{sarlin2020superglue} is trained with the indoor models in the ScanNet \cite{dai2017ScanNet} dataset and the outdoor models in the MegaDepth \cite{li2018MegaDepth} dataset.

\item \textbf{SGMNet} \cite{chen2021learning} is trained on the GL3D \cite{shen2018matchable} dataset. Our experiments utilize the SIFT version of SGMNet, where the detector and descriptor are rotation-invariant.

\item \textbf{LoFTR} \cite{sun2021loftr} is trained with the same experimental setup as SuperGlue. 

\item \textbf{MatchFormer} \cite{wang2022matchformer} is trained with the same experimental setup as SuperGlue and LoFTR. Limited by our GPU memory, the lightweight version of MatchFormer is used in our experiments.

\item \textbf{SE2-LoFTR} \cite{bokman2022case} is trained on the MegaDepth dataset.

\end{itemize}

Furthermore, the mean matching accuracy (MMA) is employed to quantify the performance of the aforementioned correspondence matching algorithms, which are run on a PC with an Intel Core i7-10870H CPU and an NVIDIA RTX3080-laptop GPU (having a 16GB DDR4 memory).

\subsection{Comparison of the SoTA approaches on the Rotated-HPatches Dataset}
\begin{figure}[t!]
	\centering
	\includegraphics[width=0.99\textwidth]{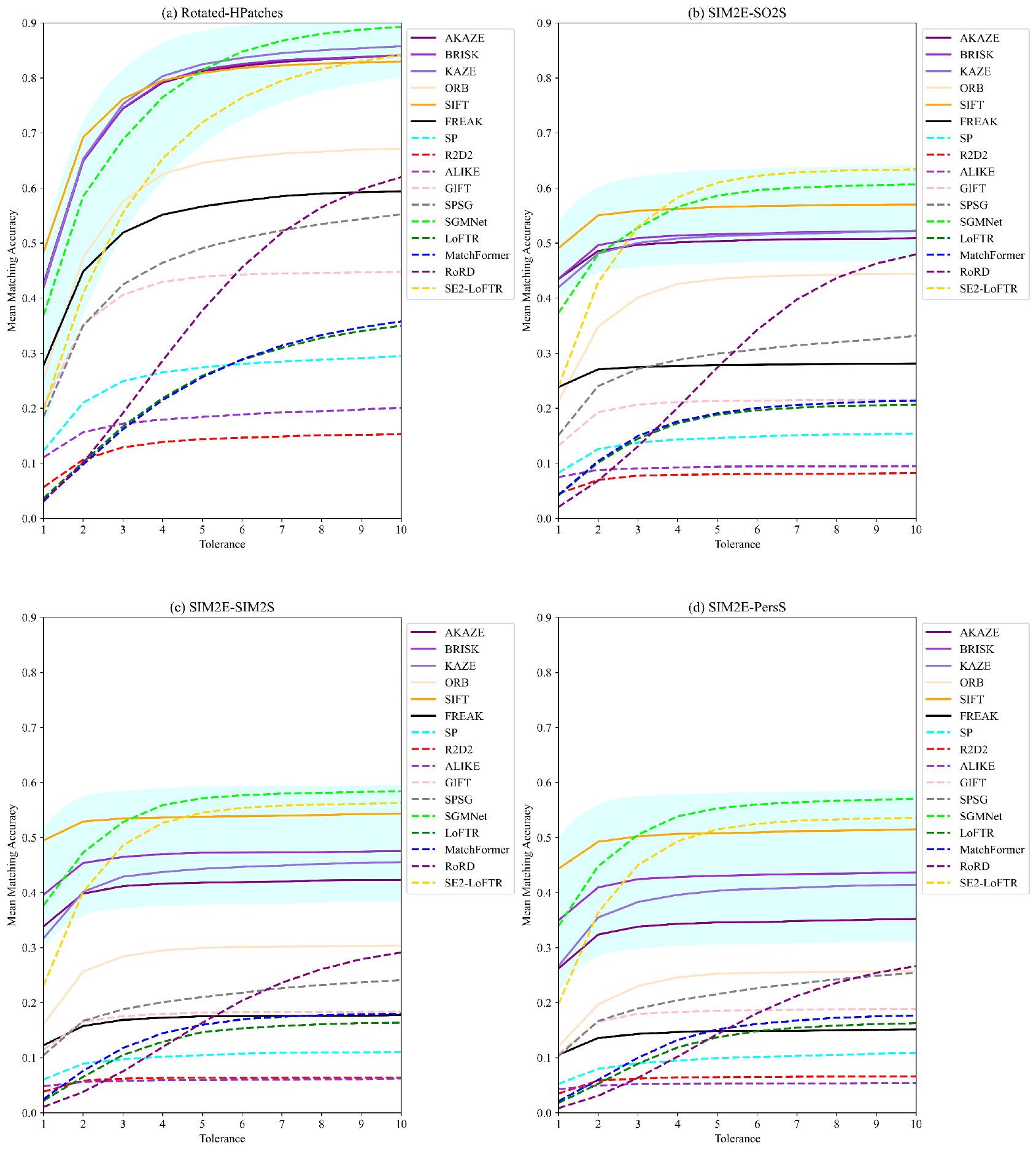}
	\caption{MMA results of six classical and ten learning-based correspondence matching approaches on the Rotated-HPatches dataset and our SIM2E-SO2S, SIM2E-SIM2S, and SIM2E-PersS subsets. MMA results for the classical methods are shown as solid lines, while MMA results for the learning-based methods are shown as dashed lines.}
	\label{fig:mmacurve}
\end{figure}

The Rotated-HPatches \cite{parihar2021rord} dataset is generated using the Hpatches \cite{balntas2017hpatches} dataset to evaluate rotation-equivariant capability of correspondence matching methods.
Each sub-folder of the Rotated-HPatches dataset contains one reference image and five target images. The target images are obtained by rotating the reference image at a random angle. The correspondence matching ground truth is acquired using the homography matrices between each pair of reference and target images. As illustrated in Fig. \ref{fig:mmacurve}(a), the SoTA correspondence matching algorithms demonstrate significantly different performances on the Rotated-HPatches dataset. 

The classical algorithms, such as AKAZE, BRISK, KAZE, and SIFT, achieve the best overall performances on the Rotated-HPatches dataset, as they consider both the scaling and rotation invariance of visual features. Benefiting from the higher dimensional feature descriptors, these four algorithms outperform ORB and FREAK. 

On the other hand, SuperPoint, R2D2, and ALIKE are developed without considering rotation invariance. Therefore, their performances are relatively poor on the Rotated-HPatches dataset. GIFT \cite{liu2019gift} uses SuperPoint as the feature detector. Its feature descriptor is developed based on G-CNN to acquire the rotation-equivariant capability. As expected, GIFT significantly outperforms SuperPoint. 

As can be seen from Table \ref{tab:rotated_hpatches} and Fig. \ref{fig:mmacurve}(a), when the tolerance $\delta$ exceeds 5, the learning-based methods demonstrate better performances than classical methods. For instance, SGMNet outperforms all classical methods when $\delta>5$ and SE2-LoFTR shows similar performance to the classical methods when $\delta>8$. Referring to \cite{chen2021learning}, SGMNet is a lightweight version of SuperGlue and demonstrates slightly worse performance than SuperGlue. However, SuperGlue with SuperPoint performs much worse than SGMNet. This is probably because SGMNet uses SIFT as its detector and descriptor, which has the rotation-equivariant capability. 

It can be observed that SoTA classical methods always perform better than learning-based methods when the tolerance is small. With the decrease in tolerance, the MMA scores achieved by learning-based methods drop considerably. This is probably because the learning-based methods are generally trained via self-supervised learning, where the tolerance to determine positive samples is typically set to 3. Furthermore, the model is difficult to converge in the training phase when the tolerance is too small, \textit{e.g.}, less than 1, while the model's accuracy degrades dramatically when the tolerance is too large. Moreover, compared to classical methods, learning-based methods can always obtain more correspondences. Therefore, improving the subpixel accuracy of learning-based approaches without reducing the number of valid matches is a research area that requires more attention.

\begin{table}[t!]
    \centering
        \caption{The performance of SoTA correspondence matching approaches on the Rotated-HPatches dataset. $N$ denotes the average number of valid matches. }
    \label{tab:rotated_hpatches}
    \setlength{\tabcolsep}{1.5mm}
    \begin{tabular}
        {l|cccc|r}
        \toprule
        Method & $\delta\leq1$ & $\delta\leq3$ & $\delta\leq5$ & $\delta\leq10$ & $N$ \\
        \hline
        AKAZE \cite{alcantarilla2011fast}  &  0.426  &  0.744  &  0.812  &  0.841 &  203 \\
        BRISK \cite{leutenegger2011brisk}  &  0.419  &  0.745  &  0.816  &  0.841 & 282 \\
        KAZE \cite{alcantarilla2012kaze} &  0.423  &  0.753  &  \textbf{0.825}  &  0.858 & 505 \\
        ORB \cite{rublee2011orb} &  0.253  &  0.576  &  0.646  &  0.672 & 20 \\
        SIFT \cite{lowe1999object} &  \textbf{0.484}  &  \textbf{0.762}  &  0.809  &  0.830 & 727 \\
        FREAK \cite{alahi2012freak} &  0.278  &  0.519  &  0.567  &  0.594 & 27 \\
        SP \cite{detone2018superpoint} &  0.123  &  0.249  &  0.274  &  0.295 & 120 \\
        R2D2 \cite{revaud2019r2d2} &  0.057  &  0.129  &  0.144  &  0.153 & 158 \\
        ALIKE \cite{zhao2022alike} &  0.111  &  0.172  &  0.184  &  0.201 & 59 \\
        GIFT \cite{liu2019gift} &  0.203  &  0.406  &  0.439  &  0.448 & 186 \\
        RoRD \cite{parihar2021rord} &  0.030  &  0.191  &  0.378  &  0.620 & 1077 \\
        SPSG \cite{sarlin2020superglue} &  0.185  &  0.425  &  0.491  &  0.552 & 479 \\
        SGMNet \cite{chen2021learning} &  0.369  &  0.688  &  0.814  &  \textbf{0.893} & 1278 \\
        LoFTR \cite{sun2021loftr} &  0.037  &  0.167  &  0.259  &  0.350 & 506 \\
        MatchFormer \cite{wang2022matchformer} &  0.033  &  0.162  &  0.256  &  0.358 & 600 \\
        SE2-LoFTR \cite{bokman2022case} &  0.197  &  0.556  &  0.720  &  0.842 & 1305 \\
		\bottomrule
    \end{tabular}
\end{table}

\begin{table}[t!]
	\centering
		\caption{The performance of SoTA correspondence matching approaches on the SIM2E-SO2S subset. $N$ denotes the average number of valid matches. }
	\label{tab:sim2e_so2s}
	\setlength{\tabcolsep}{1.5mm}
	\begin{tabular}
		{l|cccc|r}
		\toprule[1.4pt]
		Method & $\delta\leq1$ & $\delta\leq3$ & $\delta\leq5$ & $\delta\leq10$ & $N$ \\
		\midrule[1pt]
		AKAZE \cite{alcantarilla2011fast}  &  0.435  &  0.497  &  0.504  &  0.509 & 72\\
		BRISK \cite{leutenegger2011brisk}  &  0.435  &  0.509  &  0.516  &  0.522 & 113\\
		KAZE \cite{alcantarilla2012kaze}  &  0.420  &  0.500  &  0.513  &  0.522 & 129\\
		ORB \cite{rublee2011orb}  &  0.213  &  0.402  &  0.435  &  0.444 & 24\\
		SIFT \cite{lowe1999object}  &  \textbf{0.491}  &  \textbf{0.559}  &  0.566  &  0.570 & 264\\
		FREAK \cite{alahi2012freak}  &  0.238  &  0.275  &  0.278  &  0.281 & 9\\
		SP \cite{detone2018superpoint}  &  0.083  &  0.138  &  0.146  &  0.154 & 40\\
		R2D2 \cite{revaud2019r2d2}  &  0.045  &  0.078  &  0.080  &  0.082 & 53\\
		ALIKE \cite{zhao2022alike}  &  0.075  &  0.091  &  0.094  &  0.095 & 22\\
		GIFT \cite{liu2019gift}  &  0.132  &  0.207  &  0.213  &  0.215 & 51\\
		RoRD \cite{parihar2021rord}  &  0.021  &  0.131  &  0.274  &  0.480 & 630\\
		SPSG \cite{sarlin2020superglue}  &  0.151  &  0.271  &  0.299  &  0.332 & 192\\
		SGMNet \cite{chen2021learning}  &  0.373  &  0.528  &  0.586  &  0.607 & 1226\\
		LoFTR \cite{sun2021loftr}  &  0.042  &  0.145  &  0.188  &  0.206 & 329\\
		MatchFormer \cite{wang2022matchformer}  &  0.042  &  0.150  &  0.191  &  0.213 & 451\\
		SE2-LoFTR \cite{bokman2022case}  &  0.239  &  0.530  &  \textbf{0.610}  &  \textbf{0.634} & 1640\\
		\bottomrule [1.5pt]
	\end{tabular}
\end{table}

\begin{table}[t!]
	\centering
	\caption{The performance of SoTA correspondence matching approaches on the SIM2E-SIM2S subset. $N$ denotes the average number of valid matches. }
	\label{tab:sim2e_sim2s}
	\setlength{\tabcolsep}{1.5mm}
	\begin{tabular}
		{l|cccc|r}
		\toprule[1.4pt]
		Method & $\delta\leq1$ & $\delta\leq3$ & $\delta\leq5$ & $\delta\leq10$ & $N$ \\
		\midrule[1pt]
		AKAZE \cite{alcantarilla2011fast}  &  0.338  &  0.412  &  0.418  &  0.423 & 22\\
		BRISK \cite{leutenegger2011brisk}  &  0.396  &  0.465  &  0.472  &  0.475 & 51\\
		KAZE \cite{alcantarilla2012kaze}  &  0.317  &  0.429  &  0.443  &  0.455 & 54\\
		ORB \cite{rublee2011orb}  &  0.161  &  0.284  &  0.300  &  0.303 & 10\\
		SIFT \cite{lowe1999object}  &  \textbf{0.495}  &  \textbf{0.535}  &  0.538  &  0.544 & 165\\
		FREAK \cite{alahi2012freak}  &  0.123  &  0.169  &  0.175  &  0.178 & 3\\
		SP \cite{detone2018superpoint}  &  0.060  &  0.097  &  0.104  &  0.110 & 20\\
		R2D2 \cite{revaud2019r2d2}  &  0.038  &  0.062  &  0.063  &  0.064 & 23\\
		ALIKE \cite{zhao2022alike}  &  0.048  &  0.058  &  0.059  &  0.062 & 6\\
		GIFT \cite{liu2019gift}  &  0.114  &  0.175  &  0.182  &  0.184 & 32\\
		RoRD \cite{parihar2021rord}  &  0.011  &  0.075  &  0.164  &  0.291 & 287\\
		SPSG \cite{sarlin2020superglue}  &  0.105  &  0.188  &  0.210  &  0.241 & 118\\
		SGMNet \cite{chen2021learning}  &  0.377  &  0.528  &  \textbf{0.571}  &  \textbf{0.584} & 802\\
		LoFTR \cite{sun2021loftr}  &  0.022  &  0.105  &  0.146  &  0.164 & 155\\
		MatchFormer \cite{wang2022matchformer}  &  0.025  &  0.118  &  0.160  &  0.180 & 232\\
		SE2-LoFTR \cite{bokman2022case}  &  0.231  &  0.486  &  0.545  &  0.563 & 847\\
		\bottomrule [1.5pt]
	\end{tabular}
\end{table}

\begin{table}[t!]
	\centering
	\caption{The performance of SoTA correspondence matching approaches on the SIM2E-PersS subset. $N$ denotes the average number of valid matches. }
	\label{tab:sim2e_perss}
	\setlength{\tabcolsep}{1.5mm}
	\begin{tabular}
		{l|cccc|r}
		\toprule[1.4pt]
		Method & $\delta\leq1$ & $\delta\leq3$ & $\delta\leq5$ & $\delta\leq10$ & $N$ \\
		\midrule[1pt]
		AKAZE \cite{alcantarilla2011fast}  &  0.262  &  0.338  &  0.346  &  0.352 & 14\\
		BRISK \cite{leutenegger2011brisk} &  0.350  &  0.424  &  0.431  &  0.436 & 34\\
		KAZE \cite{alcantarilla2012kaze} &  0.267  &  0.383  &  0.404  &  0.414 & 42\\
		ORB \cite{rublee2011orb} &  0.120  &  0.230  &  0.253  &  0.258 & 7\\
		SIFT \cite{lowe1999object} &  \textbf{0.443}  &  0.502  &  0.508  &  0.515 & 130\\
		FREAK \cite{alahi2012freak} &  0.105  &  0.143  &  0.149  &  0.152 & 3\\
		SP \cite{detone2018superpoint} &  0.052  &  0.090  &  0.100  &  0.108 & 13\\
		R2D2 \cite{revaud2019r2d2} &  0.035  &  0.062  &  0.065  &  0.066 & 16\\
		ALIKE \cite{zhao2022alike} &  0.043  &  0.052  &  0.053  &  0.054 & 5\\
		GIFT \cite{liu2019gift} &  0.107  &  0.180  &  0.185  &  0.188 & 24\\
		RoRD \cite{parihar2021rord} &  0.008  &  0.063  &  0.143  &  0.267 & 257\\
		SPSG \cite{sarlin2020superglue} &  0.103  &  0.189  &  0.216  &  0.254 & 120\\
		SGMNet \cite{chen2021learning} &  0.339  &  \textbf{0.505}  &  \textbf{0.553}  &  \textbf{0.570} & 758\\
		LoFTR \cite{sun2021loftr} &  0.018  &  0.089  &  0.137  &  0.163 & 120\\
		MatchFormer \cite{wang2022matchformer} &  0.021  &  0.100  &  0.151  &  0.177 & 161\\
		SE2-LoFTR \cite{bokman2022case} &  0.197  &  0.450  &  0.515  &  0.536 & 772\\
		\bottomrule [1.5pt]
	\end{tabular}
\end{table}

\subsection{Comparison of the SoTA approaches on our SIM2E Dataset}

In this paper, we quantify the group-equivalent capabilities of the aforementioned algorithms on the three subsets of our SIM2E Dataset. 


\subsubsection{Experimental results on the SIM2E-SO2S subset}

Our SIM2E-SO2S subset is created in a similar fashion to the Rotated-HPatches dataset. Since we select time-lapse videos with more challenging illumination conditions, apply more uniformly distributed random rotations, and add synthetic backgrounds to the target image, the SIM2E-SO2S subset is expected to reflect the correspondence matching algorithms' group-equivariant capabilities more comprehensively. As can be observed from Fig. \ref{fig:mmacurve}(b), all the SoTA methods perform much worse on our SIM2E-SO2S subset because it is more challenging than the Rotated-HPatches dataset. Furthermore, 
similar to the experimental results in the Rotated-HPatches experiments (see Fig. \ref{fig:mmacurve}(a)), SE2-LoFTR, SGMNet, AKAZE, BRISK, KAZE, and SIFT also achieve the best group-equivariant capabilities on our SIM2E-SO2S subset (see Fig. \ref{fig:mmacurve}(b)). This validates the effectiveness of our SIM2E-SO2S subset in terms of evaluating a correspondence matching algorithm's group-equivariant capability. Moreover, the MMA scores achieved by these six methods differ more significantly in the SIM2E-SO2S experiments. Therefore, our SIM2E-SO2S subset more comprehensively quantifies the group-equivariant capability of a given correspondence matching algorithm.



\subsubsection{Experimental results on the SIM2E-SIM2S subset}

Compared to the SIM2E-SO2S subset, the SIM2E-SIM2S subset contains scaling and translation transformations that shrink the size of the overlapping area between image pairs. As illustrated in Fig. \ref{fig:mmacurve}(c) and Table \ref{tab:sim2e_sim2s}, the performances of SE2-LoFTR, SGMNet, SIFT, and BRISK remain stable in the SIM2E-SIM2S experiments, while other models' performances degrade dramatically. Therefore, our SIM2E-SIM2S subset can be used to quantify not only the rotation-equivariant capability but also the scaling-equivariant capability of correspondence matching algorithms. 

\subsubsection{Experimental results on the SIM2E-PersS subset}
Compared to the SIM2E-SO2S and SIM2E-SIM2S subsets, the SIM2E-PersS subset contains random perspective transformations. Therefore, correspondence matching on the SIM2E-PersS subset is more challenging. Similarly, in the SIM2E-PersS experiments (see Fig. \ref{fig:mmacurve}(d) and Table \ref{tab:sim2e_perss}), the performances of SE2-LoFTR, SGMNet, and SIFT remain stable, while the performances of BRISK, AKAZE, and KAZE degrade more significantly. Therefore, our SIM2E-PersS subset can be utilized to quantify sim(2)-equivariant capability of correspondence matching algorithms in a more in-depth manner.
\begin{figure}[t!]
	\centering
	\includegraphics[width=0.99\textwidth]{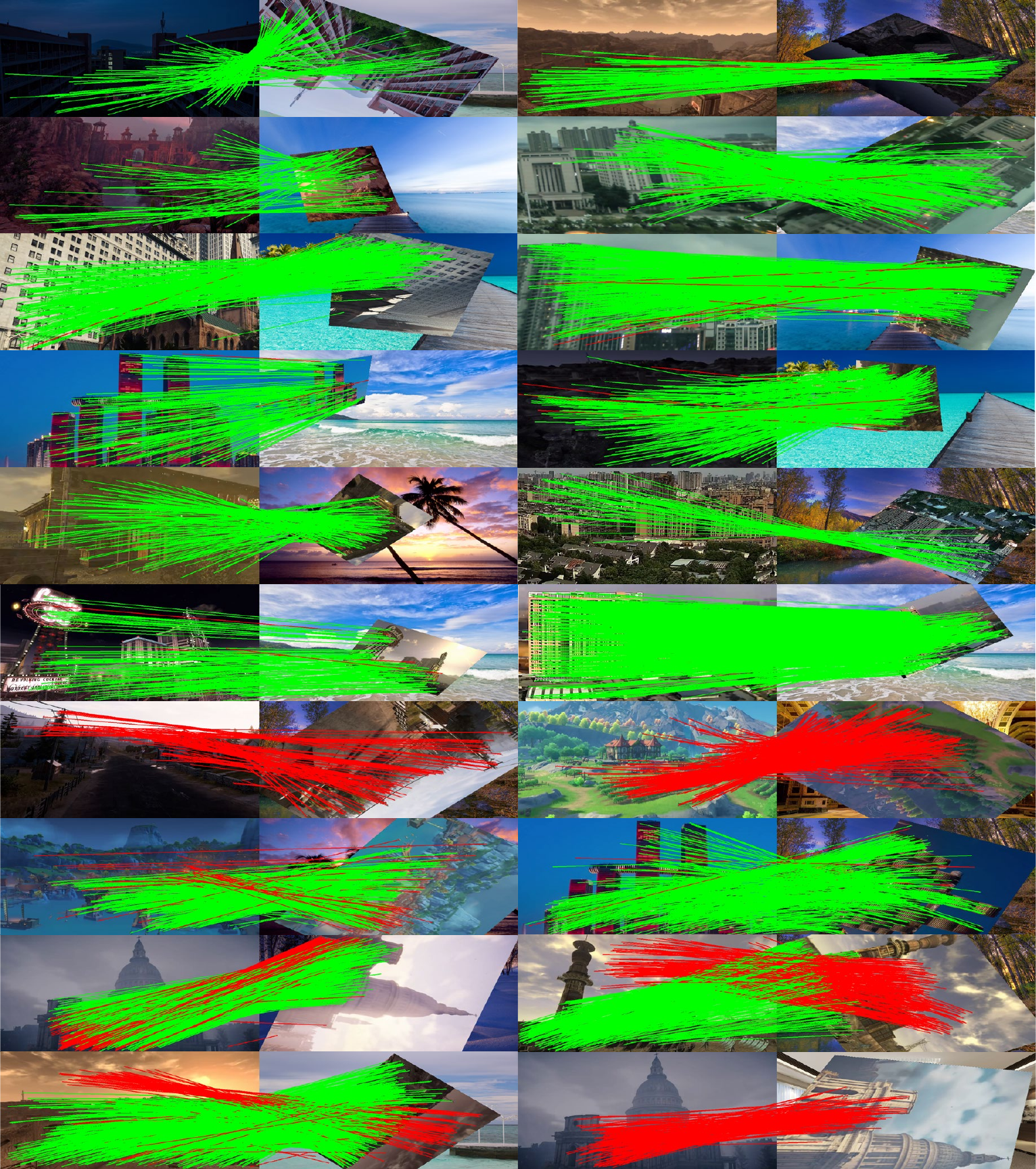}
	\caption{Correspondence matching results on our SIM2E dataset.}
	\label{fig:matchingfigure}
\end{figure}

\subsection{Discussion}
\label{sec.discussion}
The experimental results presented above show that SGMNet, SE2-LoFTR, SIFT, BRISK, KAZE, and AKAZE demonstrate similar group-equivariant capabilities. As can be seen in Table \ref{tab:rotated_hpatches}, the performances of these algorithms are very similar, while SE2-LoFTR achieves the worst performance on the Rotated-HPatches dataset. As can be observed in Fig. \ref{fig:mmacurve}(b)-(d), our created SIM2E dataset can reflect the group-equivariant capabilities of the existing correspondence matching algorithms more comprehensively. The compared SoTA methods achieve the best performances on the SIM2E-SO2S subset and the worst overall performances on the SIM2E-PersS subset. Therefore, we believe that our created SIM2E dataset can help users obtain more objective and in-depth evaluation results of their developed correspondence matching algorithms' group-equivariant capabilities.

\section{Conclusion}
\label{sec.conclusion}
This paper presented a benchmark dataset for the evaluation of sim(2)-equivariant capability of correspondence matching approaches. We first discussed the classical and learning-based methods and the mainstream of developing group-equivariant network architectures. We qualitatively and quantitatively evaluated sixteen SoTA correspondence matching algorithms on the Rotated-HPatches dataset and three subsets of our created SIM2E dataset. These results suggest that our SIM2E dataset is much more challenging than public correspondence matching datasets, and it can comprehensively reflect the group-equivariant capability of SoTA correspondence matching approaches. In summary,  group-equivariant detection, group-equivariant description, and group-equivariant position information are vital for group-equivariant correspondence matching. SuperGlue, LoFTR, and SGMNet use neural networks to fuse global position information and local feature descriptors, and achieve superior performances over others. However, obtaining group equivariance of position information is still challenging, as discussed in \cite{bokman2022case}. The scaling-equivariant and rotation-equivariant capabilities of learning-based approaches are close to classical approaches. However, the sub-pixel accuracy achieved by the former is still unsatisfactory.

\section{Acknowledgements}
	
This work was supported by the National Key R\&D Program of China, under grant No. 2020AAA0108100, awarded to Prof. Qijun Chen. This work was also supported by the Fundamental Research Funds for the Central Universities, under projects No. 22120220184, No. 22120220214, and No. 2022-5-YB-08, awarded to Prof. Rui Fan. 


\bibliographystyle{unsrt}

\begin{thebibliography}{10}
	
	\bibitem{zhou2009object}
	Huiyu Zhou et~al.
	\newblock Object tracking using {SIFT} features and mean shift.
	\newblock {\em Computer vision and image understanding}, 113(3):345--352, 2009.
	
	\bibitem{yu2022accurate}
	Yang Yu et~al.
	\newblock Accurate and robust visual localization system in large-scale
	appearance-changing environments.
	\newblock {\em IEEE/ASME Transactions on Mechatronics}, 2022.
	\newblock DOI: 10.1109/TMECH.2022.3177237.
	
	\bibitem{ling2016high}
	Yonggen Ling and Shaojie Shen.
	\newblock High-precision online markerless stereo extrinsic calibration.
	\newblock In {\em 2016 IEEE/RSJ International Conference on Intelligent Robots
		and Systems (IROS)}, pages 1771--1778. IEEE, 2016.
	
	\bibitem{fan2018road}
	Rui Fan et~al.
	\newblock Road surface {3D} reconstruction based on dense subpixel disparity
	map estimation.
	\newblock {\em IEEE Transactions on Image Processing}, 27(6):3025--3035, 2018.
	
	\bibitem{brown2007automatic}
	Matthew Brown and David~G Lowe.
	\newblock Automatic panoramic image stitching using invariant features.
	\newblock {\em International journal of computer vision}, 74(1):59--73, 2007.
	
	\bibitem{fan2019road}
	Rui Fan and Ming Liu.
	\newblock Road damage detection based on unsupervised disparity map
	segmentation.
	\newblock {\em IEEE Transactions on Intelligent Transportation Systems},
	21(11):4906--4911, 2019.
	
	\bibitem{moravec1977techniques}
	Hans~P Moravec.
	\newblock Techniques towards automatic visual obstacle avoidance.
	\newblock 1977.
	
	\bibitem{harris1988combined}
	Chris Harris, Mike Stephens, et~al.
	\newblock A combined corner and edge detector.
	\newblock In {\em Alvey vision conference}, volume~15, pages 10--5244.
	Citeseer, 1988.
	
	\bibitem{lowe1999object}
	David~G Lowe.
	\newblock Object recognition from local scale-invariant features.
	\newblock In {\em Proceedings of the seventh IEEE international conference on
		computer vision}, volume~2, pages 1150--1157. Ieee, 1999.
	
	\bibitem{morel2009asift}
	Jean-Michel Morel and Guoshen Yu.
	\newblock {ASIFT}: A new framework for fully affine invariant image comparison.
	\newblock {\em SIAM journal on imaging sciences}, 2(2):438--469, 2009.
	
	\bibitem{rosten2006machine}
	Edward Rosten and Tom Drummond.
	\newblock Machine learning for high-speed corner detection.
	\newblock In {\em European conference on computer vision}, pages 430--443.
	Springer, 2006.
	
	\bibitem{calonder2010brief}
	Michael Calonder, Vincent Lepetit, Christoph Strecha, and Pascal Fua.
	\newblock {BRIEF}: Binary robust independent elementary features.
	\newblock In {\em European conference on computer vision}, pages 778--792.
	Springer, 2010.
	
	\bibitem{rublee2011orb}
	Ethan Rublee, Vincent Rabaud, Kurt Konolige, and Gary Bradski.
	\newblock {ORB}: An efficient alternative to {SIFT} or {SURF}.
	\newblock In {\em 2011 International conference on computer vision}, pages
	2564--2571. Ieee, 2011.
	
	\bibitem{mur2015orb}
	Raul Mur-Artal, Jose Maria~Martinez Montiel, and Juan~D Tardos.
	\newblock {ORB-SLAM}: A versatile and accurate monocular {SLAM} system.
	\newblock {\em IEEE transactions on robotics}, 31(5):1147--1163, 2015.
	
	\bibitem{mur2017orb}
	Raul Mur-Artal and Juan~D Tard{\'o}s.
	\newblock {ORB-SLAM2}: An open-source {SLAM} system for monocular, stereo, and
	{RGB-D} cameras.
	\newblock {\em IEEE transactions on robotics}, 33(5):1255--1262, 2017.
	
	\bibitem{yi2016lift}
	Kwang~Moo Yi, Eduard Trulls, Vincent Lepetit, and Pascal Fua.
	\newblock {LIFT}: Learned invariant feature transform.
	\newblock In {\em European conference on computer vision}, pages 467--483.
	Springer, 2016.
	
	\bibitem{detone2018superpoint}
	Daniel DeTone, Tomasz Malisiewicz, and Andrew Rabinovich.
	\newblock {SuperPoint}: Self-supervised interest point detection and
	description.
	\newblock In {\em Proceedings of the IEEE conference on computer vision and
		pattern recognition workshops}, pages 224--236, 2018.
	
	\bibitem{lin2014microsoft}
	Tsung-Yi Lin, Michael Maire, Serge Belongie, James Hays, Pietro Perona, Deva
	Ramanan, Piotr Doll{\'a}r, and C~Lawrence Zitnick.
	\newblock Microsoft {COCO}: Common objects in context.
	\newblock In {\em European conference on computer vision}, pages 740--755.
	Springer, 2014.
	
	\bibitem{dusmanu2019d2}
	Mihai Dusmanu, Ignacio Rocco, Tomas Pajdla, Marc Pollefeys, Josef Sivic,
	Akihiko Torii, and Torsten Sattler.
	\newblock {D2-Net}: A trainable cnn for joint description and detection of
	local features.
	\newblock In {\em Proceedings of the IEEE/cvf conference on computer vision and
		pattern recognition}, pages 8092--8101, 2019.
	
	\bibitem{revaud2019r2d2}
	Jerome Revaud, Philippe Weinzaepfel, C{\'e}sar De~Souza, Noe Pion, Gabriela
	Csurka, Yohann Cabon, and Martin Humenberger.
	\newblock {R2D2}: Repeatable and reliable detector and descriptor.
	\newblock {\em arXiv preprint arXiv:1906.06195}, 2019.
	
	\bibitem{tyszkiewicz2020disk}
	Micha{\l} Tyszkiewicz, Pascal Fua, and Eduard Trulls.
	\newblock {DISK}: Learning local features with policy gradient.
	\newblock {\em Advances in Neural Information Processing Systems},
	33:14254--14265, 2020.
	
	\bibitem{sarlin2020superglue}
	Paul-Edouard Sarlin, Daniel DeTone, Tomasz Malisiewicz, and Andrew Rabinovich.
	\newblock {SuperGlue}: Learning feature matching with graph neural networks.
	\newblock In {\em Proceedings of the IEEE/CVF conference on computer vision and
		pattern recognition}, pages 4938--4947, 2020.
	
	\bibitem{cuturi2013sinkhorn}
	Marco Cuturi.
	\newblock Sinkhorn distances: Lightspeed computation of optimal transport.
	\newblock {\em Advances in neural information processing systems}, 26, 2013.
	
	\bibitem{chen2021learning}
	Hongkai Chen, Zixin Luo, Jiahui Zhang, Lei Zhou, Xuyang Bai, Zeyu Hu, Chiew-Lan
	Tai, and Long Quan.
	\newblock Learning to match features with seeded graph matching network.
	\newblock In {\em Proceedings of the IEEE/CVF International Conference on
		Computer Vision}, pages 6301--6310, 2021.
	
	\bibitem{sun2021loftr}
	Jiaming Sun, Zehong Shen, Yuang Wang, Hujun Bao, and Xiaowei Zhou.
	\newblock {LoFTR}: Detector-free local feature matching with transformers.
	\newblock In {\em Proceedings of the IEEE/CVF conference on computer vision and
		pattern recognition}, pages 8922--8931, 2021.
	
	\bibitem{wang2022matchformer}
	Qing Wang, Jiaming Zhang, Kailun Yang, Kunyu Peng, and Rainer Stiefelhagen.
	\newblock {MatchFormer}: Interleaving attention in transformers for feature
	matching.
	\newblock {\em arXiv preprint arXiv:2203.09645}, 2022.
	
	\bibitem{parihar2021rord}
	Udit~Singh Parihar, Aniket Gujarathi, Kinal Mehta, Satyajit Tourani, Sourav
	Garg, Michael Milford, and K~Madhava Krishna.
	\newblock {RoRD}: Rotation-robust descriptors and orthographic views for local
	feature matching.
	\newblock In {\em 2021 IEEE/RSJ International Conference on Intelligent Robots
		and Systems (IROS)}, pages 1593--1600. IEEE, 2021.
	
	\bibitem{cohen2016group}
	Taco Cohen and Max Welling.
	\newblock Group equivariant convolutional networks.
	\newblock In {\em International conference on machine learning}, pages
	2990--2999. PMLR, 2016.
	
	\bibitem{li2018deep}
	Junying Li, Zichen Yang, Haifeng Liu, and Deng Cai.
	\newblock Deep rotation equivariant network.
	\newblock {\em Neurocomputing}, 290:26--33, 2018.
	
	\bibitem{cohen2018spherical}
	Taco~S Cohen, Mario Geiger, Jonas K{\"o}hler, and Max Welling.
	\newblock Spherical {CNNs}.
	\newblock {\em arXiv preprint arXiv:1801.10130}, 2018.
	
	\bibitem{weiler2019general}
	Maurice Weiler and Gabriele Cesa.
	\newblock General {E(2)}-equivariant steerable {CNNs}.
	\newblock {\em Advances in Neural Information Processing Systems}, 32, 2019.
	
	\bibitem{liu2019gift}
	Yuan Liu, Zehong Shen, Zhixuan Lin, Sida Peng, Hujun Bao, and Xiaowei Zhou.
	\newblock {GIFT}: Learning transformation-invariant dense visual descriptors
	via group {CNNs}.
	\newblock {\em Advances in Neural Information Processing Systems}, 32, 2019.
	
	\bibitem{lee2022self}
	Jongmin Lee, Byungjin Kim, and Minsu Cho.
	\newblock Self-supervised equivariant learning for oriented keypoint detection.
	\newblock {\em arXiv preprint arXiv:2204.08613}, 2022.
	
	\bibitem{peri2022ref}
	Abhishek Peri, Kinal Mehta, Avneesh Mishra, Michael Milford, Sourav Garg, and
	K~Madhava Krishna.
	\newblock {ReF}-rotation equivariant features for local feature matching.
	\newblock {\em arXiv preprint arXiv:2203.05206}, 2022.
	
	\bibitem{bokman2022case}
	Georg B{\"o}kman and Fredrik Kahl.
	\newblock A case for using rotation invariant features in state of the art
	feature matchers.
	\newblock {\em arXiv preprint arXiv:2204.10144}, 2022.
	
	\bibitem{cieslewski2018matching}
	Titus Cieslewski, Michael Bloesch, and Davide Scaramuzza.
	\newblock Matching features without descriptors: implicitly matched interest
	points.
	\newblock {\em arXiv preprint arXiv:1811.10681}, 2018.
	
	\bibitem{geiger2012we}
	Andreas Geiger, Philip Lenz, and Raquel Urtasun.
	\newblock Are we ready for autonomous driving? the {KITTI} vision benchmark
	suite.
	\newblock In {\em 2012 IEEE conference on computer vision and pattern
		recognition}, pages 3354--3361. IEEE, 2012.
	
	\bibitem{bokman2021zz}
	Georg B{\"o}kman, Fredrik Kahl, and Axel Flinth.
	\newblock {ZZ-Net}: A universal rotation equivariant architecture for {2D}
	point clouds.
	\newblock {\em arXiv preprint arXiv:2111.15341}, 2021.
	
	\bibitem{sattler2012image}
	Torsten Sattler, Tobias Weyand, Bastian Leibe, and Leif Kobbelt.
	\newblock Image retrieval for image-based localization revisited.
	\newblock In {\em BMVC}, volume~1, page~4, 2012.
	
	\bibitem{sattler2018benchmarking}
	Torsten Sattler, Will Maddern, Carl Toft, Akihiko Torii, Lars Hammarstrand,
	Erik Stenborg, Daniel Safari, Masatoshi Okutomi, Marc Pollefeys, Josef Sivic,
	et~al.
	\newblock Benchmarking 6{DOF} outdoor visual localization in changing
	conditions.
	\newblock In {\em Proceedings of the IEEE conference on computer vision and
		pattern recognition}, pages 8601--8610, 2018.
	
	\bibitem{dai2017ScanNet}
	Angela Dai, Angel~X Chang, Manolis Savva, Maciej Halber, Thomas Funkhouser, and
	Matthias Nie{\ss}ner.
	\newblock {ScanNet}: Richly-annotated {3D} reconstructions of indoor scenes.
	\newblock In {\em Proceedings of the IEEE conference on computer vision and
		pattern recognition}, pages 5828--5839, 2017.
	
	\bibitem{li2018MegaDepth}
	Zhengqi Li and Noah Snavely.
	\newblock {MegaDepth}: Learning single-view depth prediction from internet
	photos.
	\newblock In {\em Proceedings of the IEEE Conference on Computer Vision and
		Pattern Recognition}, pages 2041--2050, 2018.
	
	\bibitem{taira2018inloc}
	Hajime Taira, Masatoshi Okutomi, Torsten Sattler, Mircea Cimpoi, Marc
	Pollefeys, Josef Sivic, Tomas Pajdla, and Akihiko Torii.
	\newblock {InLoc}: Indoor visual localization with dense matching and view
	synthesis.
	\newblock In {\em Proceedings of the IEEE Conference on Computer Vision and
		Pattern Recognition}, pages 7199--7209, 2018.
	
	\bibitem{wang2020tartanair}
	Wenshan Wang, Delong Zhu, Xiangwei Wang, Yaoyu Hu, Yuheng Qiu, Chen Wang, Yafei
	Hu, Ashish Kapoor, and Sebastian Scherer.
	\newblock {TartanAir}: A dataset to push the limits of visual {SLAM}.
	\newblock In {\em 2020 IEEE/RSJ International Conference on Intelligent Robots
		and Systems (IROS)}, pages 4909--4916. IEEE, 2020.
	
	\bibitem{balntas2017hpatches}
	Vassileios Balntas, Karel Lenc, Andrea Vedaldi, and Krystian Mikolajczyk.
	\newblock {HPatches}: A benchmark and evaluation of handcrafted and learned
	local descriptors.
	\newblock In {\em Proceedings of the IEEE conference on computer vision and
		pattern recognition}, pages 5173--5182, 2017.
	
	\bibitem{bradski2000opencv}
	Gary Bradski.
	\newblock The {OpenCV} library.
	\newblock {\em Dr. Dobb's Journal: Software Tools for the Professional
		Programmer}, 25(11):120--123, 2000.
	
	\bibitem{alcantarilla2011fast}
	Pablo~F Alcantarilla and T~Solutions.
	\newblock Fast explicit diffusion for accelerated features in nonlinear scale
	spaces.
	\newblock {\em IEEE Trans. Patt. Anal. Mach. Intell}, 34(7):1281--1298, 2011.
	
	\bibitem{leutenegger2011brisk}
	Stefan Leutenegger, Margarita Chli, and Roland~Y Siegwart.
	\newblock {BRISK}: Binary robust invariant scalable keypoints.
	\newblock In {\em 2011 International conference on computer vision}, pages
	2548--2555. Ieee, 2011.
	
	\bibitem{alcantarilla2012kaze}
	Pablo~Fern{\'a}ndez Alcantarilla, Adrien Bartoli, and Andrew~J Davison.
	\newblock {KAZE} features.
	\newblock In {\em European conference on computer vision}, pages 214--227.
	Springer, 2012.
	
	\bibitem{alahi2012freak}
	Alexandre Alahi, Raphael Ortiz, and Pierre Vandergheynst.
	\newblock {FREAK}: Fast retina keypoint.
	\newblock In {\em 2012 IEEE conference on computer vision and pattern
		recognition}, pages 510--517. Ieee, 2012.
	
	\bibitem{radenovic2018revisiting}
	Filip Radenovi{\'c}, Ahmet Iscen, Giorgos Tolias, Yannis Avrithis, and
	Ond{\v{r}}ej Chum.
	\newblock Revisiting oxford and paris: Large-scale image retrieval
	benchmarking.
	\newblock In {\em Proceedings of the IEEE conference on computer vision and
		pattern recognition}, pages 5706--5715, 2018.
	
	\bibitem{zhao2022alike}
	Xiaoming Zhao, Xingming Wu, Jinyu Miao, Weihai Chen, Peter~CY Chen, and
	Zhengguo Li.
	\newblock {ALIKE}: Accurate and lightweight keypoint detection and descriptor
	extraction.
	\newblock {\em IEEE Transactions on Multimedia}, 2022.
	
	\bibitem{shen2018matchable}
	Tianwei Shen, Zixin Luo, Lei Zhou, Runze Zhang, Siyu Zhu, Tian Fang, and Long
	Quan.
	\newblock Matchable image retrieval by learning from surface reconstruction.
	\newblock In {\em Asian conference on computer vision}, pages 415--431.
	Springer, 2018.
	
	\bibitem{snavely2006photo}
	Noah Snavely, Steven~M Seitz, and Richard Szeliski.
	\newblock Photo tourism: Exploring photo collections in {3D}.
	\newblock In {\em ACM siggraph 2006 papers}, pages 835--846. 2006.
	
\end{thebibliography}

\end{document}